# SIMILARITY MATCHING TECHNIQUES FOR FAULT DIAGNOSIS IN AUTOMOTIVE INFOTAINMENT ELECTRONICS

**Dr. Mashud Kabir**

**Department of Computer Science, University of Tuebingen**
**D-72027 Tuebingen, Germany**
***mashud98@yahoo.com***

**Abstract**

Fault diagnosis has become a very important area of research during the last decade due to the advancement of mechanical and electrical systems in industries. The automobile is a crucial field where fault diagnosis is given a special attention. Due to the increasing complexity and newly added features in vehicles, a comprehensive study has to be performed in order to achieve an appropriate diagnosis model. A diagnosis system is capable of identifying the faults of a system by investigating the observable effects (or symptoms). The system categorizes the fault into a diagnosis class and identifies a probable cause based on the supplied fault symptoms. Fault categorization and identification are done using similarity matching techniques. The development of diagnosis classes is done by making use of previous experience, knowledge or information within an application area. The necessary information used may come from several sources of knowledge, such as from system analysis. In this paper similarity matching techniques for fault diagnosis in automotive infotainment applications are discussed.

***Key words:*** *similarity, fault, diagnosis, matching, automotive, infotainment, cosine.*

## 1. Introduction

At first feature selection is discussed where stop word list and word stemming are used. Then pattern recognition is explained. Ranking algorithms are used to rank words, web pages. Page ranking algorithm is discussed keeping in mind our application. Similarity algorithms are discussed in the next sections. Then proper similarity matching algorithm which best fits to fault diagnosis in automotive infotainment system is presented. The algorithm is analyzed with real field data and the results are evaluated.

## 2. Feature Selection

Feature selection is one of the main steps in similarity matching of faults. We apply a stop word [1] list to filter out the meaningless words. A list of stop words has been built. This list has been created keeping in mind the existing standard fault description language in automotive infotainment systems.

Word stemming is a method where lexically similar words are listed together. Here, the words with affixes and suffixes are converted into root words. This methodology overcomes the limitation of words with the same meaning being categorized into different classes. A list of stemming words is created for automotive infotainment system. Both the lists of stop words and stemming words were developed with the help of experienced system engineers in automotive infotainment system.

## 3. Pattern Recognition

Based on highly developed skill after sensing the surroundings, humans are capable of taking any actions according to their observations. By observing the nature of human intelligence, a machine can be built to do the same job, such as identifying hand writing, post code, voice, finger print, DNA, human face etc.

A pattern is an abstract object such as a set of measurements describing a physical object. This is an entity with a given name such as hand writing, a sentence, human face etc. Pattern recognition consists of several steps such as observation of inputs, learning how to distinguish different patterns and making rationale decisions in categorizing patterns.

Shmuel Brody [3] has summarized the concepts of pattern recognition and their uses in similarity matching. Human detected patterns contain many relevant and irrelevant data. The most important task in pattern recognition is to find out the meaningful patterns and to disregard the irrelevant subject matter. The fields of area of pattern recognition range from data analysis, feature extraction, error estimation, error removal, cluster analysis, grammatical inference and parsing.

Faramarz Valafar [4] has discussed pattern recognition techniques in data analysis. Clustering is one of the most commonly used recognition techniques. Data are grouped into clusters or groups in clustering. K-means clustering [2] is a widely used algorithm for data clustering. In k-means similar algorithm patterns are partitioned into the same group. All the data are classified into any of the k clusters or classes. Then the mean inter and intra-class distances are determined. The last step is to maximize the intra-class distance and minimize the inter-class distance. This is an iterative procedure where data is moved from one cluster to another. This process continues until optimized distances of intra-class and inter-class are found.

In pattern recognition different techniques are applied for similarity matching. For this work it is necessary to discover optimized techniques and algorithms for similarity matching, fault classifying and fault cause detecting.

## 4. Ranking Algorithms

PageRank algorithm is a widely used algorithm to rank web pages according to their importance. The algorithm is described as following –

PageRank is a link analysis algorithm to rank a web page from a set of pages according to its relative importance. It provides a numerical weighting to each of the page elements in the set. This weighting is called PageRank of E which is denoted by PR(E).

PageRank was introduced by Larry Page at Stanford University to develop a new search engine in the web. The ranking of a page depends on the number of links of the other pages to that page.

PageRank is a probability distribution which shows the likelihood that a user randomly clicking on the links finds a specific site. This probability ranges from 0 to 1. A PageRank of 0.8 means that the probability of reaching a specific site by randomly clicking on a set of links is 80%.

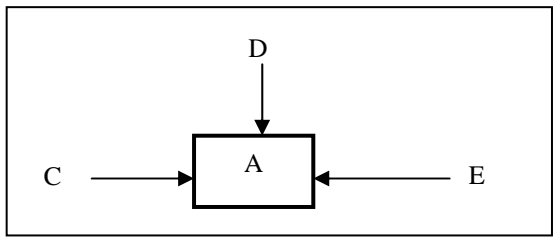

Fig. 1  Inbound link of page A.

A set of five web pages is assumed: A, B, C, D, E. The initial probability is distributed evenly among these pages. Therefore, each of the pages will get a PageRank of 1.0/5. It means,
PR(A) = PR(B) = PR(C) = PR(D) = PR(E) = 0.2     (i)

Now suppose the scenario as depicted in figure 1:

Page A has inbound links from Page C, D and E. Thus, the PageRank of page A

$$PR(A) = PR(C) + PR(D) + PR(E) \qquad (ii)$$

Page C has other outbound links to page E, page D has other outbound links to B, C and E as depicted in figure 2.

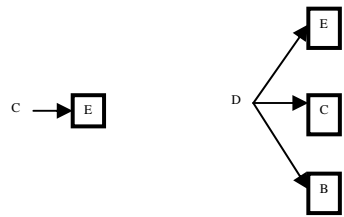

Fig. 2  The Outbound links of page C and Page D.

The value of the link-votes is divided among all the outbound links of a page. Thus, page C contributes a vote weight of 0.2/2 i.e. 0.1 and page D contributes a vote weight of 0.2/4 i.e. 0.05.

Thus, the equation stands in the following form:

$$PR(A) = \frac{PR(C)}{2} + \frac{PR(D)}{4} + \frac{PR(E)}{1} \qquad (iii)$$

The above equation can be generalized in the following form assuming that the PageRank incurred by an outband link of a page is the page's own PageRank in the set divided by the number of outband links

$$PR(A) = \frac{PR(C)}{L(C)} + \frac{PR(D)}{L(D)} + \frac{PR(E)}{L(E)} \qquad (iv)$$

The PageRank of any page $i$ can be expressed in the following form:

$$PR(i) = \sum_{j \varepsilon S_i} \frac{PR(j)}{N_j} \qquad (v)$$

Where,

$PR(i)$ = PageRank of page i

$PR(j)$ = PageRank of any other pages except page i.

$N_j$ = Number of pages in the set

$j \varepsilon S_i$ = Inbound pages linking to page *i*

PageRank algorithm is mainly used for internet applications to find the rank of a page. The basis of the algorithm is that the rank of a page depends on the inbound links of the other pages. To apply this technique we need to compare links among the pages with the links among the features of the fault. But this study requires the ranking of features according to their importance. This makes PageRank algorithm inappropriate for this project.

## 5. Similarity Matching

This chapter describes the similarity matching techniques for strings. Using these techniques, a concept is proposed to search similar faults when the symptoms of a fault are provided.

Edit distance is a common term in matching algorithms. The word distance is used to compare different data for similarity. Edit distance is a measure to estimate differences between input elements. Different methods to calculate edit distance exist:

**Levenshtein Distance**

Levenshtein distance is named after the Russian scientist Vladimir Levenshtein, who devised the algorithm in 1965. The Levenshtein distance between two strings is given by the minimum number of operations needed to transform one string into the other, where an operation is an insertion, deletion, or substitution of a single character.

Levenshtein distance (LD) is a measure of the similarity between two inputs: the source *s* and the target input *t*. The distance is the number of deletions, insertions, or substitutions required to transform *s* into *t*. For example,
If s is "math" and t is "math", then LD(s,t) = 0, because no transformations are needed.
If s is "math" and t is "mats", then LD(s,t) = 1, because one substitution (change "h" to "s") is sufficient to transform s into t.
The more different the inputs are, the greater the Levenshtein distance is.

Insertion, deletion and substitution are the main criteria for determining Levenshtein Distance. The position of a character plays an important role to determine the distance. In this study, the description of a fault is dealt with. If Levenshtein Distance is applied to find out the similarity of faults it would not give a meaningful result as the positions of the strings should not have importance. That is why this technique will not be used in this study.

**Damerau-Levenshtein Distance**

Damerau-Levenshtein distance comes from Levenshtein distance that counts transposition as a single edit operation. The Damerau-Levenshtein distance is equal to the minimal number of insertions, deletions, substitutions and transpositions needed to transform one string into the other.

Kukich [5] described several edit distance algorithms which use Damerau-Levenshtein distance. It has been proved that the use of Damerau-Levenshtein metric to calculate the similarity between two words is a slow process. For this reason this method is not well-suited for similarity matching in this project.

**Needleman – Wunsch Distance**

The Levenshtein distance algorithm assumes that the cost of all insertions, deletions, substitutions or conversions is equal. However, in some scenarios this may not be desirable and may mask the acceptable distances between inputs.

Needleman-Wunsch has modified Levenshtein distance algorithm to add cost matrix as an extra input. This matrix structure contains two cost matrics for each pair of characters to convert from and to. The cost of inserting this character and converting between characters is listed in this matrix.

This approach is not appropriate for use in this study's similarity matching for the same reason stated in Levenshtein approach.

**Hamming Distance**

The Hamming distance [6] H is defined for the same length inputs. For two inputs *s* and *t*, H(s, t) is the number of places in which the two strings differ, i.e., have different characters.

Hamming Distance is used in information theory. This method can not be applied in similarity matching for automotive faults since Hamming Distance only considers the differences among the two inputs.

**Weighted Edit Distance**

This algorithm differs from the Edit Distance in weighting. A particular weight is imposed for each operation of insertion, deletion and substitution.

The main goal of similarity matching of faults is to find the faults with the similar behaviors. Weighted Edit Distance focuses on providing weight on the operations. This kind of approach is inappropriate for finding similar faults.

**Hamming Distance**

The Hamming distance is the number of positions for which the corresponding characters differ. It is simply the number of differences between two strings of the same length. For example:
The Hamming Distance between GERMANY and IRELAND is 5.

To apply this distance between two error features they must be of equal length, which is a rare case. This results in the decision not to use Hamming Distance for similarity matching in this study.

## 6. Similarity Determination

The aim of this section is to propose an algorithm to use for similarity matching in text queries. The procedures of this algorithm are as following

A text (query) T is represented by multidimensional vector:
$F(T) = (F_1(T), F_2(T), …F_k(T))$ (occurrence vector)
k = no. of distinct term occurring in database (non-stop word)

Function of frequency of the *i*-th term in T,

$$F_i(T) = \frac{1}{2}\left(1 + \frac{tf_i^T}{\max tf_i^T}\right)\log\frac{N}{n_i}$$

where,
$tf_i^T$ = frequency of the *i-th* term in ***T***
$\max tf_i^T$ = no. of database documents where the most frequent term of T occurs
$N$ = no. of database entries
$n_i$ = no. of entries where the *i-th* term occurs

The cosine similarity measure between a query (*A*) and a stored document (*B*) is defined as:

$$Cos_{similarity}(A, B) = \frac{\sum_{i=1}^{N} \alpha_i F_i(A) F_i(B)}{\sqrt{\sum_{i=1}^{N} \alpha_i^2 F_i(A)^2 \sum_{i=1}^{N} \alpha_i^2 F_i(B)^2}}$$

where,
$\alpha_i$ = user-determined parameter (weights) (*~1*)

Cosine similarity method counts the number of different words in two documents. With this method the highest frequency words within any document will have the largest influence on its similarity with other documents. Documents with many occurrences of an unusual word or many different unusual words will have low cosine similarity measures with most other documents. Weighting schemes are frequently used to modify the standard cosine measure. These typically lower the importance of common words.

Below are the results of some input data and their similarities with the existing input database using this algorithm –

**Input Database:**

This is the database which is already stored in the system. This is compared with the user provided fault symptoms.

| Attachment | Defect ID | Fault Characteristics |
|---|---|---|
| | 32 | *Display ON Signal will be sent, but Display remains dark* |
| Y | 40 | *Preconditions: radio hu message* |
| | 41 | *radio: radio. message; audio hu radio message message* |
| | 42 | *Preconditions: message-> hu -> audio* |
| Y | 44 | *preconditions: hu sds sdars message message radio radio* |
| | 45 | *radio hu message message message* |
| | 46 | *radio hu no message message* |
| | 47 | *radio no hu message message no sds* |
| | 48 | *radio hu message message no message sds* |
| | 49 | *radio hu* |
| | 50 | *radio* |
| | 51 | *radio radio* |
| | 52 | *radio radio radio* |
| | 53 | *radio does not receive message from headunit* |

**Result Analysis:**

Below is the graphical representation of outputs for determining fault similarities corresponding to user provided fault symptoms. The similarities of fault symptom radio hu are 100% (fault id 49) with the database fault radio hu and 68% (fault id 40) with the database fault Preconditions: radio hu message. The result of this fault matching is shown in figure 3.

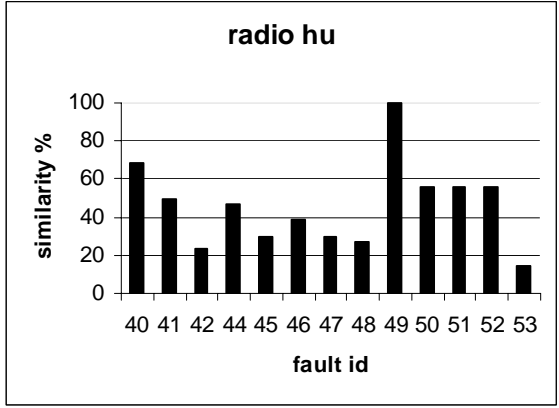

Fig. 3  Fault similarities with symptom radio hu.

The similarities of fault symptom radio hu message are 100% (id 40) with database fault radio hu and 84% (id 41) with database fault radio: radio. message; audio hu radio message message and 84% (id 45) with database fault radio hu message message messaeg. The result of this fault matching is shown in figure 4.

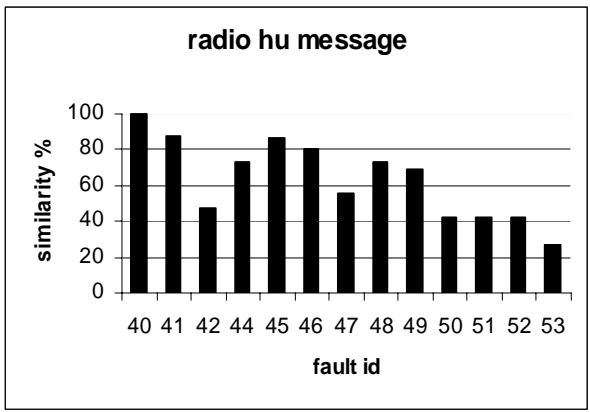

Fig. 4  Fault similarities with symptom radio hu message.

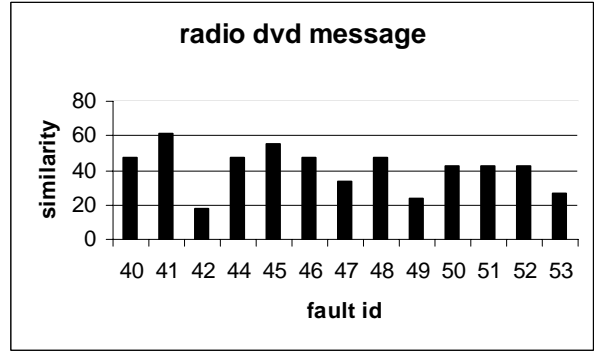

Fig. 5  Fault similarities with symptom radio dvd message.

The similarities of fault symptom radio dvd message are 61% (id 41) with the database fault radio: radio. message; audio hu radio message message and 56% (id 45) with the database fault radio hu message message message. The result of this fault matching is shown in figure 5.

The similarities of fault symptom radio dvd are 58% with database faults radio (id 50) and radio radio(id 51) and radio radio radio (id 52). The result of this fault matching is shown in figure 6.

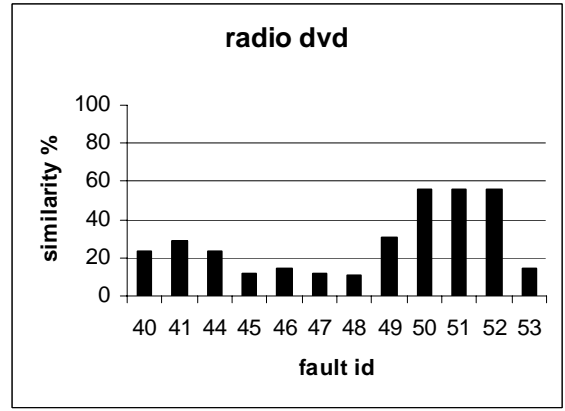

Fig. 6  Fault similarities with symptom radio dvd.

Based on the above result analysis it can be concluded that the similarity of a user provided fault is higher if the symptom of the fault matches more closely with any database fault. It satisfies the requirement of finding similar faults for a fault symptom.

## 7. Conclusion

In this paper feature selection, pattern recognition, page ranking algorithms have been discussed to process input data and system database. Different similarity matching algorithms have also been explained. Cosine similarity algorithm has been chosen for our special application of automotive infotainment system. Real field automotive faults have been used to analyse the cosine similarity method. After comparing with the existing fault database, a decision of fault similarities on user provided fault has been made and the results have been discussed.

**Mashud Kabir.** I was born in Narayanganj, Bangladesh in 1976. I have completed my Bachelor of Science (BSc) in Electrical & Electronic Engineering from Bangladesh University of Engineering & Technology (BUET) in 2000. I was awarded board scholarship from 1995 to 2000. I earned my Master of Science (MSc) in Communication Engineering from University of Stuttgart, Germany in 2003. I achieved STIEBET German Government scholarship during my Master Study. My Master thesis was "Region-Based Adaptation of Diffusion Protocols in MANETs" where up to 21% of broadcast can be saved. I worked at Mercedes-Benz Technology Center, Germany from 2003 to 2005 as a PhD student. I have worked in the research & development projects of BMW, Land-Rover and Audi in Automotive Infotainment Network area for more than four years. I have achieved my Doctoral degree from the Department of Computer Science, University of Tuebingen, Germany in 2008. My dissertation topic was "Intelligent System for Fault Diagnosis in Automotive Applications".